\documentclass{article}

\PassOptionsToPackage{square}{natbib}
\usepackage{natbib}
\setcitestyle{numbers,square}

\usepackage[preprint]{neurips_2023}

\usepackage[utf8]{inputenc} 
\usepackage[T1]{fontenc}    
\usepackage{url}            
\usepackage{booktabs}       
\usepackage{multirow}
\usepackage{amsfonts}       
\usepackage{nicefrac}       
\usepackage{microtype}      
\usepackage{xcolor}         
\usepackage{bbding}
\usepackage{ulem}
\usepackage{pifont}
\usepackage{graphicx}
\usepackage[colorlinks,linkcolor=blue, anchorcolor=blue, citecolor=blue]{hyperref}
\usepackage{color}
\usepackage{colortbl}

\definecolor{mygray}{gray}{.9}

\title{COSA: Concatenated Sample Pretrained  Vision-Language Foundation Model}

\newcommand*{\affaddr}[1]{#1} 
\newcommand*{\affmark}[1][*]{\textsuperscript{#1}}
\newcommand*\samethanks[1][\value{footnote}]{\footnotemark[#1]}

\author{%
Sihan Chen\affmark[1]\affmark[2]\thanks{Work done during an internship at Bytedance.}, Xingjian He\affmark[2], Handong Li\affmark[1]\affmark[2], Xiaojie Jin\affmark[3]\thanks{Corresponding authors.}, Jiashi Feng\affmark[3]\samethanks[2], Jing Liu\affmark[1]\affmark[2]\samethanks[2]\\
\normalsize
\affaddr{\affmark[1]School of Artificial Intelligence, University of Chinese Academy of Sciences} \\
\affaddr{\affmark[2]
Institute of Automation, Chinese Academy of Science}
\affaddr{\affmark[3]Bytedance Inc}\\
\normalsize
{\tt \{sihan.chen, xingjian.he, jliu\}@nlpr.ia.ac.cn}, {\tt \ lihandong2023@ia.ac.cn} \\
{\tt \{jinxiaojie,jshfeng\}@bytedance.com}
}

\begin{document}

\maketitle

\begin{abstract}

  Due to the limited scale and quality of video-text training corpus, most  vision-language  foundation  models employ  image-text datasets for pretraining and primarily focus on modeling visually semantic representations while disregarding temporal semantic representations and correlations. To address this issue, we propose COSA, a \textbf{CO}ncatenated \textbf{SA}mple pretrained vision-language foundation model. COSA jointly models visual contents and event-level temporal cues using only image-text corpora.  We achieve this by sequentially concatenating multiple image-text pairs as inputs for pretraining. This transformation effectively converts existing image-text corpora into a pseudo long-form video-paragraph corpus, enabling richer scene transformations and explicit event-description correspondence. Extensive experiments demonstrate that COSA consistently improves performance across a broad range of downstream tasks, including long-form/short-form video-text tasks and image-text tasks such as retrieval, captioning, and question answering. Notably, COSA achieves state-of-the-art results on various competitive benchmarks. Code and model are released at \url{https://github.com/TXH-mercury/COSA}.

\end{abstract}

\section{Introduction}

Image-text and video-text pretraining models have garnered increasing attention in recent years due to their strong capabilities in bridging the domains of vision and language. These models contribute significantly to various vision-language tasks such as cross-modality retrieval, vision captioning, and visual question answering (VQA). In the early stages, these models were independently trained using distinct cross-modality corpora, network architectures, and training objectives. However, considering that a video can be perceived as a sequence of multiple images~\cite{lei2021less}, and an image can be seen as a frozen video~\cite{bain2021frozen}, most researchers lean towards training unified foundation models using corpora from both domains, employing sparsely sampled frames as the video representation.

Nevertheless, the majority of large-scale vision-language foundation models are usually trained solely on image-text corpora, disregarding the joint modeling of images and videos. This limitation stems from the restricted scale of video-text corpora, which is associated with the costly processes of video uploading, storage, and downloading. Currently, widely accessible web-crawled video-text corpora, such as WebVid10M~~\cite{bain2021frozen}, are still two orders of magnitude smaller than its image-text counterpart (LAION-2B~\cite{schuhmann2021laion}), which implies the dominance of image-text data in joint modeling. On the other hand, image-text pretraining can also bring benefits to video-related tasks. In fact, some large-scale image-text foundation models~~\cite{radford2021learning, wang2022git} have achieved competitive or even superior performance on video tasks compared to dedicated video-language pretraining models.

Despite the fact that image-text foundation models have acquired a broad range of semantic correlations between visual concepts and language, they totally overlook temporal contexts within videos, which are crucial for video-language reasoning. 
This raises the question that \textit{whether we can utilize large-scale image-text corpora to train a unified foundation model capable of capturing both static and temporal information during the pretraining phase.}

\begin{figure*}[t]
\centering
\includegraphics[width=1.0\linewidth]{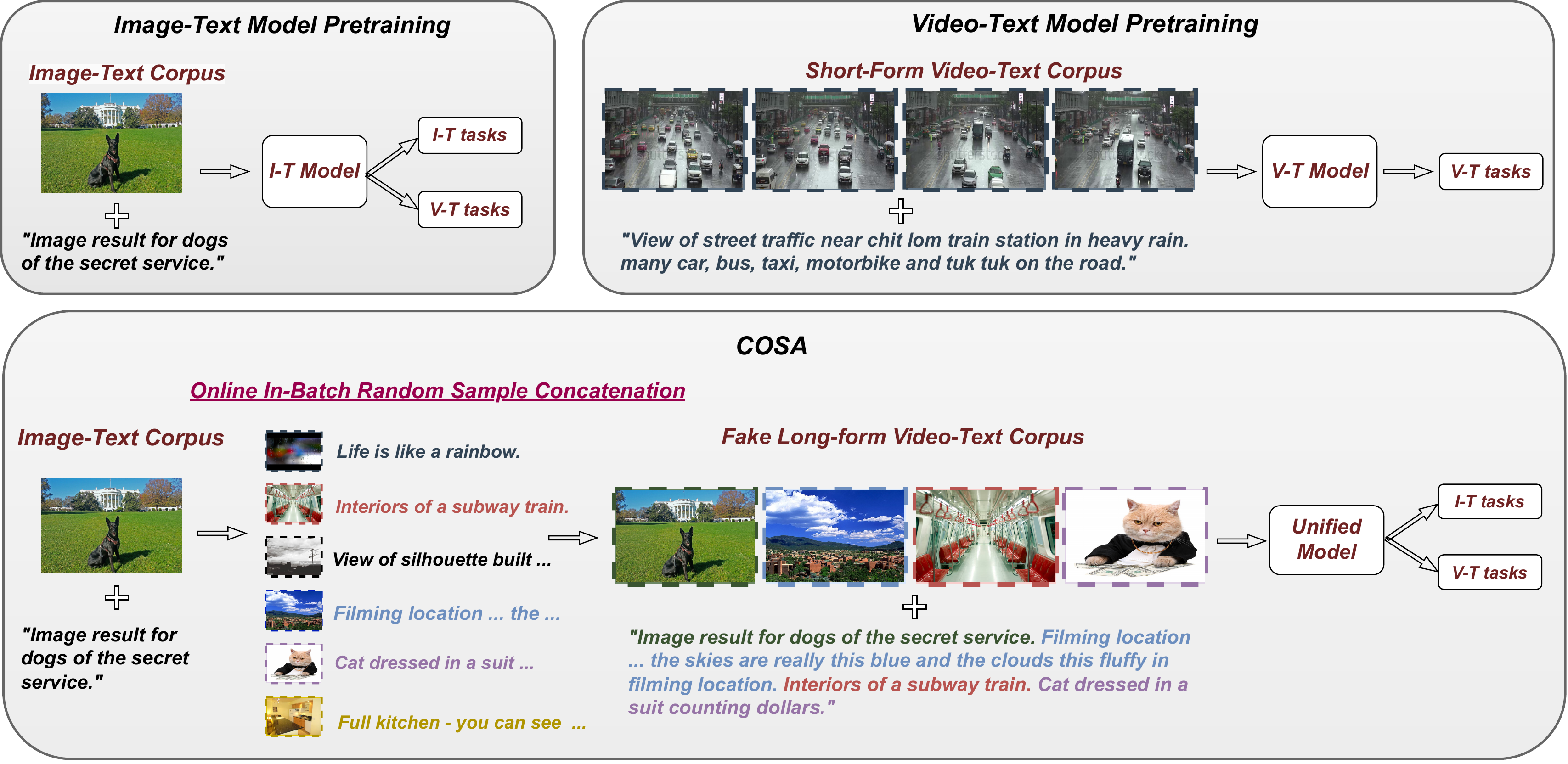}
\caption{Visualizations of the traditional image-text and video-text model pretraining pipeline and the proposed unified COSA, which transforms the image-text corpus into a synthetic long-form video-text corpus online through random sample concatenation.}
\label{fig:data representaion}
\end{figure*}

Towards this goal, we propose a simple yet effective approach: training a foundation model with concatenated samples. By dynamically transforming the image-text corpus into a long-form video-paragraph corpus, we enhance the learning of both image-language and video-language representations. Specifically, at each training step, for every image-text training sample, we randomly select a certain number of samples from the same batch and concatenate them together. As illustrated in Figure \ref{fig:data representaion}, the concatenation process operates on both images and texts, ensuring an explicit correspondence between events and sentences. In comparison to short-form video-text corpora~\cite{bain2021frozen}, the on-the-fly concatenated corpus offers richer scene transformations, reduced vision redundancy resulting from sampling randomness, and longer fine-grained captions that describe every frame in sequential order. Moreover, compared to long-form video-text corpora such as HowTo100M~\cite{miech2019howto100m} or HD\_VILA\_100M~\cite{xue2022advancing}, the composed corpus provides more accurate descriptions related to visual content, rather than relying on noisy automatic speech recognition (ASR) transcriptions that possess weak correlations with visual content.
 
Enhanced by sample concatenation, we deliver a unified vision-language foundation model called COSA. COSA possesses a simple architecture and takes concatenated samples as input during pretraining. This model is capable of handling both discriminative and generative tasks, including visual retrieval, captioning, and question answering. Extensive ablation studies have demonstrated that sample concatenation is effective for both image-text and video-text corpora, consistently improving the performance of downstream tasks in long-form and short-form video-language as well as image-language scenarios. We attribute this improvement to the fact that concatenated samples contain a wealth of information compared to individual samples, thereby aiding the model in learning better vision-language correlations and temporal correspondence within the multiple images and texts context. Through extensive experiments, COSA achieves new state-of-the-art results across a wide range of benchmarks and various model scale settings.

\section{Related Work}

\textbf{Temporal Learning in Video-Language Pretraining (VidLP).} Temporal information is the essence that distinguishes videos from images. Early methods aimed to enhance temporal learning in VidLP by either employing models pretrained on action classification tasks as vision backbones~\cite{bain2021frozen, li2022align, fu2021violet, fu2022empirical} or designing specific training objectives such as frame order modeling (FOM)~\cite{li2020hero, zellers2021merlot} or masked video modeling (MVM)~\cite{fu2021violet, fu2022empirical}. However, Singularity~\cite{lei2022revealing} reveals that temporal modeling in pretraining may not be necessary, as sampling a single frame can yield comparable performance to utilizing multiple frames. They attribute this to the static appearance bias in video-language downstream benchmarks. We hypothesize that another contributing factor is the inadequacy of the current video-text corpus (WebVid-2.5M~\cite{bain2021frozen}) for both action-level and event-level temporal learning. In terms of action-level temporal learning, the corpus contains fewer annotated actions compared to recognition datasets like Kinetics-400~\cite{kay2017kinetics}, and there is significant caption redundancy as actions are often associated with several predicates rather than the entire sentence. For event-level temporal learning, the corpus is typically constrained to a single scene and description, whereas rich scene transformations and event-sentence correspondence are crucial. To address these limitations, OmniVL~\cite{wang2022omnivl} extends the idea of UniCL~\cite{lu2022unified} to the video domain and incorporates Kinetics-400 into the training corpus. HiTeA~\cite{ye2022hitea} tackles caption redundancy through specific fine-grained alignment designs. LF\_VILA~\cite{sun2022long} constructs a long-form dataset by leveraging the large-scale HD\_VILA\_100M~\cite{xue2022advancing} dataset to enhance event-level temporal learning. In contrast to the aforementioned methods, we enhance event-level temporal learning through concatenating samples, which dynamically composes a synthetic long-form training corpus online based on existing image-text or short-form video-corpus datasets, resulting in improved performance.

\textbf{Unified Vision-Language Foundation Model.} The Frozen model~\cite{bain2021frozen} was the first to leverage a mixture of image-text (CC3M~\cite{sharma2018conceptual}) and video-text corpus (WebVid2.5M~\cite{bain2021frozen}) for training VidLP models by sparsely sampling the video corpus and treating the image corpus as frozen videos. Subsequent VidLP method have followed this approach, supporting both image-text and video-text tasks, with advanced vision backbones~\cite{fu2021violet, fu2022empirical, ye2022hitea} or specially designed training objectives~\cite{li2022align, ye2022hitea, ge2022miles}.

Regarding large-scale unified vision-language foundation models, they can be classified into two types: co-training methods and two-stage training methods. Co-training methods train on both image and video corpora simultaneously and typical representatives of co-training methods is Flamingo~\cite{alayrac2022flamingo}. Two-stage training methods involve initially training a large model on image-text corpora and then adapting it for video-text tasks. In contrast to their video-text counterparts, image-text foundation models can be scaled up more economically in terms of both data size and parameter size, thanks to the abundance of web data. They can be trained using contrastive learning~\cite{radford2021learning, jia2021scaling}, generative learning~\cite{wang2022git, chen2022pali, wang2021simvlm}, or a combination of both~\cite{yu2022coca}. For adapting to video-text tasks, CLIP-based models~\cite{luo2022clip4clip, xue2022clip} employ temporal fusion transformers and frame-word fine-grained alignment learning to enable text-to-video retrieval. GIT~\cite{wang2022git} directly concatenates multiple frame features as input to the text decoder and achieves state-of-the-art performance on most video captioning benchmarks. VideoCoCa~\cite{yan2022video} adapts the CoCa model~\cite{yu2022coca} through continuous learning on video-language corpora. MaMMUT~\cite{kuo2023mammut} applies a similar temporal perceiving capability enhancement during finetuning as TubeViT~\cite{piergiovanni2022rethinking}. In comparison to the above methods, COSA simultaneously model image-text and video-text learning during the pretraining stage,  with image-text corpora utilized only.

 \begin{figure*}[t]
\centering
\includegraphics[width=1.0\linewidth]{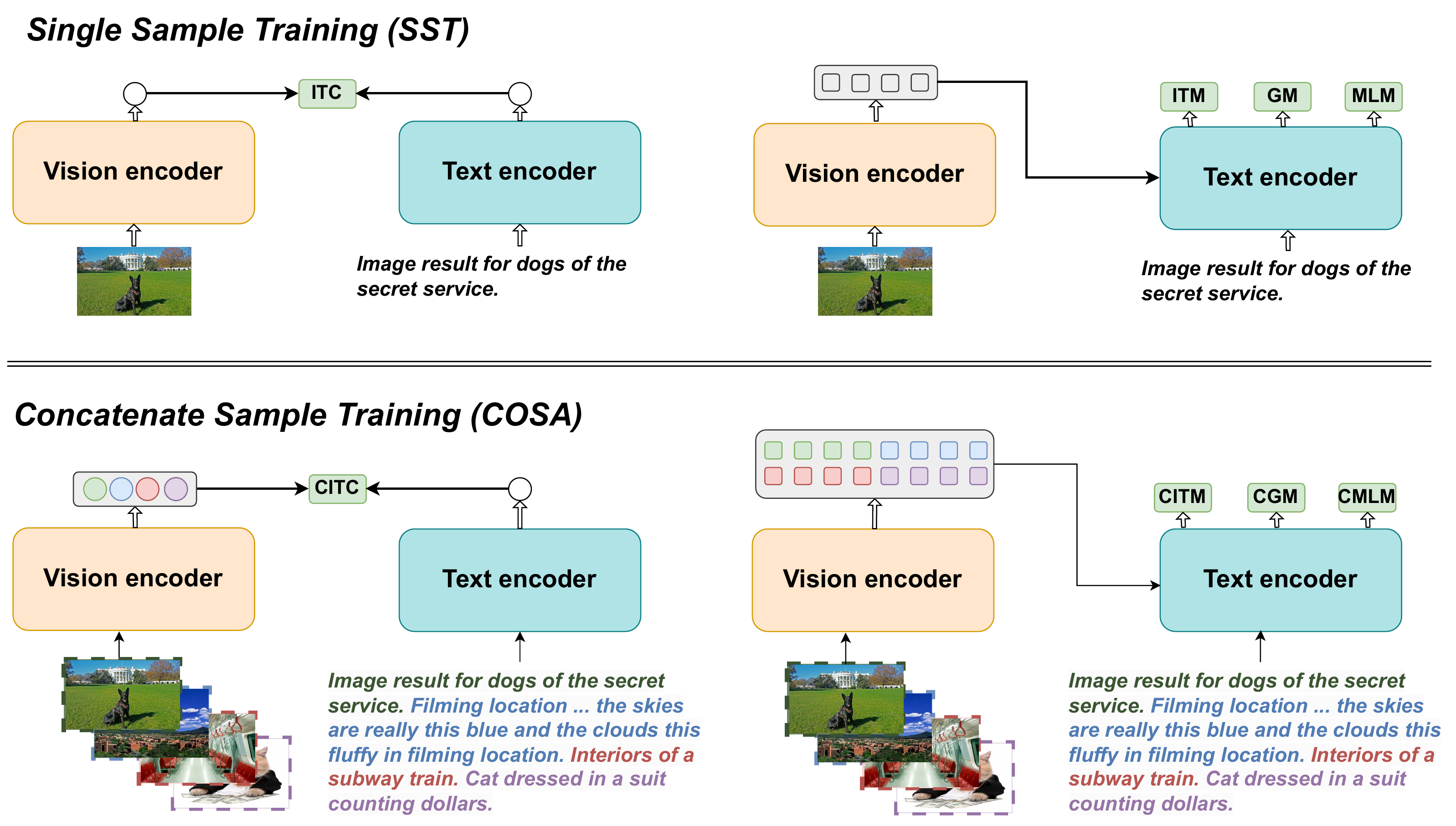}
\caption{Visualizations of the training framework for COSA (bottom). In contrast to the conventional single sample training framework (SST), COSA  takes the on-the-fly transformed pseudo long-form video-paragraph corpus as input. Circles and squares in the figure represent global and patch features, respectively.}
\label{fig:model}
\end{figure*}

\section{Method}

\subsection{Simple Model Architecture}
COSA adopts a simple architecture, consisting of a vision encoder (ViT~\cite{dosovitskiy2020image}) and a text encoder (BERT~\cite{devlin2018bert}), which facilitates both single-modality encoding and cross-modality interaction. Specifically, we introduce additional cross-attention layers between the self-attention and feed-forward layers of BERT. These cross-attention layers are activated during the forward pass for cross-modality understanding or generation, while they are deactivated during the forward pass for single-modality encoding. By incorporating this design, the text encoder can serve as a single-modality encoder, cross-modality encoder, and decoder.

\subsection{Concatenated Sample Training}
\textbf{Online Sample Concatenation.}
As illustrated in Figure \ref{fig:model}, unlike traditional foundation models that take a single image-text pair or frames-text pair as input, COSA employs an online transformation to convert them into a pseudo long-form video-paragraph corpus. To illustrate this process, let's consider the image-text corpus. Given a batch containing $n_{b}$ image-text samples, for the $i$-th sample $S_{i}$ with its image and text denoted as $I_{i}$ and $T_{i}$ respectively, we randomly select $n_{c}$ ($n_{c}=3$ by default) other samples from the current batch and group them together. Consequently, the single sample $(I_{i}, T_{i})$ is transformed into a grouped sample $(I_{group}=[I_{i},I_{j1},...,I_{jn_{c}}], T_{group}=[T_{i},T_{j1},...,T_{jn_{c}}])$. Subsequently, we concatenate both the images and texts in the same order to establish their temporal correspondence. Specifically, for the grouped images, we individually feed them into the vision encoder, incorporate learnable temporal embeddings into the output features, and concatenate them along the sequential dimension. Regarding the grouped texts, we directly concatenate them into a long paragraph where the temporal relations can be encoded by BERT's vanilla position embedding layer. Through the process of in-order sample concatenation and the use of position embeddings, the model can learn explicit scene-level temporal alignment. This concatenation procedure is applied to all samples in the batch. COSA can also be applied on video-text corpus and in this scenario, we sample a single frame from each video and perform the same concatenation process.

The concatenated images can be viewed as a series of snapshots from multiple clips within a pseudo long-form video, where each clip captures a different scene accompanied by its corresponding descriptions. It is important to note that these pseudo videos exhibit rapid scene changes, and the clips are independent of each other due to the random sampling process. This approach effectively reduces redundancy but sacrifices clip relevance. However, our experiments demonstrate that the cost of sacrificing clip relevance appears negligible, and random sampling yields superior results compared to other settings where clip relevance is intentionally enforced.

\textbf{Training Objectives.}
We employ four training objectives, namely Image-Text Contrastive (ITC), Image-Text Matching (ITM), Masked Language Modeling (MLM), and Generative Modeling (GM), to enhance the model's cross-modality alignment, understanding, and generation capabilities. These objectives are modified to accommodate the use of concatenated samples as input, as opposed to single samples.

\textit{Concatenated Image-Text Contrastive (CITC).}
The pseudo video frames are individually passed through the vision encoder, and the average pooling of the output [CLS] token features serves as the global representation of the pseudo videos. The [CLS] token feature from BERT is utilized as the paragraph representation. A bi-directional contrastive loss is employed to bring paired concatenated samples closer together while pushing away unpaired ones.

\textit{Concatenated Image-Text Matching (CITM).}
This task requires the model to determine whether a long-form video and a paragraph correspond to each other. Hard negative mining is used following the approach in~\cite{li2021align}. The [CLS] token feature from the multimodal encoder (text encoder) is used to compute a binary classification loss through a two-layer MLP.

\textit{Concatenated Masked Language Modeling (CMLM).}
We randomly mask 15\% of the tokens in the paragraphs and employ BERT's vanilla prediction layer to reconstruct the masked tokens given the context of the pseudo long-form videos.

\textit{Concatenated Generation Modeling (CGM).}
We randomly mask 60\% of the tokens in the paragraphs and utilize the same prediction layer as in CMLM to reconstruct the masked tokens in the context of the pseudo videos. CGM incorporates causal attention masks in the self-attention layers of BERT to prevent information leakage and enhance the model's capabilities for text generation.

In addition to the four training objectives that utilize concatenated samples as input, we also include two vanilla objectives (ITC and ITM) that take raw single samples as input to enhance the model's ability to process single samples.

\begin{table}[t]
\caption{Model configurations of COSA. CC14M is a combination of CC3M~\cite{sharma2018conceptual}, CC12M~\cite{changpinyo2021conceptual}, COCO~\cite{lin2014microsoft}, SBU~\cite{ordonez2011im2text}, and VG~\cite{krishna2017visual} datasets. The largest 1.2B COSA model is trained with a data bootstrapping strategy inspired by BLIP~\cite{li2022blip}, and  captions in training corpora (marked with `*') are generated by us through an additionally trained captioner, more details can be found in Appendix. The 400M image-text pairs used in CLIP and LAION-400M~\cite{schuhmann2021laion} used in EVAClip are also included in the training examples statistics. LAION-102M is a randomly sampled subset from LAION-400M.}
\centering
\label{model_configurations}
\scalebox{0.7}{
\begin{tabular}{llllll}
\toprule
Model      & Vision Encoder   & Param & Training corpora        & Example & Step \\
\midrule
COSA-B  & Swin-B~\cite{liu2021swin}    &  251M          & CC3M+WebVid2.5M                      & 5M     & 100K   \\
COSA-B  & Swin-B~\cite{liu2021swin}    &   251M         & CC14M+WebVid2.5M                      & 17M     & 160K   \\
COSA-L & CLIP/ViT-L/14~\cite{radford2021learning}    & 468M           & CC14M+WebVid2.5M            & 417M    & 160K  \\
COSA       & EVAClip/VIT-G/14~\cite{sun2023eva} &      1.2B      & CC14M*+LAION-102M* &    415M    &   150K    \\
\bottomrule
\end{tabular}}
\end{table}

\begin{table}[]
\caption{Performance comparison on text-to-video retrieval benchmarks. For fair comparison, evaluation results before employing post-processing methods such as  dual softmax~\cite{cheng2021improving} are compared. Recall@1,5,10 are used as evaluation metrics.}

\centering
\label{sota_ret}
\scalebox{0.65}{
\begin{tabular}{ll|ccc|ccc|ccc|ccc}
\toprule
\multirow{2}{*}{Method}                                     & \multirow{2}{*}{Example} & \multicolumn{3}{c}{MSRVTT} & \multicolumn{3}{c}{DiDeMo} & \multicolumn{3}{c}{LSMDC} & \multicolumn{3}{c}{ActivityNet} \\
\cmidrule (lr){3-5} \cmidrule (lr){6-8} \cmidrule (lr){9-11} \cmidrule (lr){12-14}
                                                            &                            & R@1     & R@5     & R@10   & R@1     & R@5     & R@10   & R@1     & R@5    & R@10   & R@1       & R@5      & R@10     \\
\midrule

ClipBert~\cite{lei2021less}                                                    & 5.4M                       & 22.0    & 46.8    & 59.9   & 20.4    & 48.0    & 60.8   &    -     &      -  &     -   &       21.3    &      49.0     &  63.5         \\
Frozen~\cite{bain2021frozen}                                                      & 5M                       & 31.0     & 59.5   & 70.5    & 34.6     & 65.0     &74.7   & 15.0    & 30.8   & 39.8   &    -       &    -      &        -  \\
BridgeFormer~\cite{ge2022bridging}                                                & 5M                       & 37.6    & 64.8    & 75.1   & 37.0    & 62.2    & 73.9   & 17.9    & 35.4   & 44.5   &      -     &      -    &     -     \\
MILES~\cite{ge2022miles}                                                      & 5M                       & 37.7    & 63.6    & 73.8   & 36.6    & 63.9    & 74.0   & 17.8    & 35.6   & 44.1   &     -      &    -      &       -   \\
OA-Trans~\cite{wang2022object}                                                   & 5M                       & 35.8    & 63.4    & 76.5   & 34.8    & 64.4    & 75.1   & 18.2    & 34.3   & 43.7   &     -      &     -     &    -      \\
Clover~\cite{huang2022clover}                                                     & 5M                       & 38.6    & 67.4    & 76.4   & 45.1    & 74.3    & 82.2   & 22.7    & 42.0   & 52.6   &      -     &       -   & -         \\
VIOLETv2~\cite{fu2022empirical} &5M &37.2&64.8& 75.8 & 47.9&76.5& 84.1& 24.0& 43.5 & 54.1&-&-&-   \\
LF-VILA~\cite{sun2022long}  & 8.5M                       &     -    &     -    &    -    & 35.0    & 64.5    & 75.8   &   -      &    -    &      -  & 35.3      & 65.4     &      -    \\
\rowcolor{mygray}
\textbf{COSA-B(Ours)}   & \textbf{5M}   & \textbf{42.2}&\textbf{69.0} &\textbf{79.0} & \textbf{57.8}& \textbf{80.6}& \textbf{87.9}& \textbf{27.3}& \textbf{46.7}&\textbf{55.2} &\textbf{55.6}&\textbf{80.7} &\textbf{89.1} \\

\midrule
OmniVL~\cite{wang2022omnivl}                                                      & 17M                        & \textbf{47.8}    & \textbf{74.2}    & \textbf{83.8}   & 52.4    & 79.5    & 85.4   &  -       &      -  &   -     & -          &     -     &   -       \\

HiTeA~\cite{ye2022hitea}                                                       & 17M                        & 46.8    & 71.2    & 81.9   & 56.5    & 81.7    & 89.7   & 28.7    & 50.3   & \textbf{59.0}   & 49.7      & 77.1     & 86.7     \\
SINGULARITY~\cite{lei2022revealing}                                                & 17M                        & 41.5    & 68.7    & 77.0   & 53.9    & 79.4    & 86.9   &  -       &    -    &     -   & 47.1      & 75.5     & 85.5     \\

VINDLU-B~\cite{cheng2022vindlu} & 17M & 45.3 & 69.9 &79.6 & 59.2 &84.1   & 89.5&- & -&- &54.4&  80.7 & 89.0 \\

LAVENDER~\cite{li2022lavender}                                                   & 30M                        & 40.7    & 66.9    & 77.6   & 53.4    & 78.6    & 85.3   & 26.1    & 46.4   & 57.3   &       -    &     -     &  -        \\
\rowcolor{mygray}
\textbf{COSA-B(Ours)}   & \textbf{17M}  & 46.9 & 72.1 & 81.3& \textbf{64.1}& \textbf{86.1} & \textbf{90.6}& \textbf{31.2}  & \textbf{50.9} & 57.8 & \textbf{59.3}&  \textbf{83.8} & \textbf{90.9}\\

\midrule
All-in-one~\cite{wang2022all}  &        138M                    & 37.9    & 68.1    & 77.1   & 32.7    & 61.4    & 73.5   & 22.4    & 53.7   & 67.7   &  -         & -         &  -        \\
CLIP4Clip~\cite{luo2022clip4clip}                                                  & 400M                       & 44.5    & 71.4    & 81.6   & 43.4    & 70.2    & 80.6   & 22.6    & 41.0   & 49.1   & 40.5      & 72.4     &   -       \\
X-CLIP~\cite{ma2022x}                                                     & 400M                       & 49.3    & 75.8    & 84.8   & 47.8    & 79.3    &   -     & 26.1    & 48.4   & 46.7   & 46.2      & 75.5     &      -    \\

mPLUG-2~\cite{xu2023mplug}                                                     & 417M                       & 53.1    & 77.6    & 84.7   & 56.4    & 79.1    & 85.2   & 34.4    & 55.2   & 65.1   &        -   &    -      &     -     \\

UMT-L~\cite{li2023unmasked}                                                       & 425M                       & \textbf{58.8}    & \textbf{81.0}    & \textbf{87.1}   & 70.4    & \textbf{90.1}    & \textbf{93.5}   & \textbf{43.0}    & \textbf{65.5}   & \textbf{73.0}   & 66.8      & \textbf{89.1}     & 94.9     \\
VALOR-L~\cite{chen2023valor}                                                  & 433.5M                     & 54.4    & 79.8    & 87.6  & 57.6   & 83.3    & 88.8   & 31.8    & 52.8   &62.4   & 63.4      & 87.8     & 94.1    \\
CLIP-VIP~\cite{xue2022clip}                                                  & 500M                       & 54.2    & 77.2    & 84.8   & 50.5    & 78.4    & 87.1   & 29.4    & 50.6   & 59.0   & 53.4      & 81.4     & 90.0     \\

 \rowcolor{mygray}
\textbf{COSA-L(Ours)}                                                      & \textbf{417M}    & 54.4    &  77.2        &    84.7     &   68.3     &       88.1   &  91.7        &    38.6     &    59.1      & 67.4       &   66.8      &  87.6         &           93.9     \\

\rowcolor{mygray}
\textbf{COSA(Ours)}                                                       &   \textbf{415M}                         &  57.9 
        &  79.6      &    86.1       &    \textbf{70.5}     &     89.3     &     92.4     &       39.4  &      60.4   &  67.7       &   \textbf{67.3}        &   89.0        &    \textbf{95.0}      \\
\bottomrule
\end{tabular}}
\end{table}

\section{Experiments}

\subsection{Implementation Details}
We train COSA models using the PyTorch framework and 64 Tesla V100 cards. To ensure a fair comparison with state-of-the-art video-language pretraining models and vision-language foundation models, we train four model variants with different parameter and data sizes, as illustrated in Table \ref{model_configurations}. All models utilize BERT-Base as the text encoder. The initial learning rate is set to 1e-4, and a 10\% warm-up strategy with a linear decay schedule is employed. The batch size is set to 2048. For ablation studies, we utilize the frozen CLIP/ViT-B/16~\cite{radford2021learning} as the vision encoder for efficiency purposes, and the models are trained on CC3M for 30K steps with a batch size of 1024, unless specified otherwise.

\begin{table*}[]
\caption{Performance comparison on open-ended video QA benchmarks. Accuracy is used as the evaluation metric.}
\label{sota_qa}
\centering
\scalebox{0.65}{
\begin{tabular}{ll|cccc}
\toprule
Method        & Example  & MSRVTT-QA        & MSVD-QA          & TGIF-QA    & ActivityNet-QA  \\
\midrule
ClipBERT~\cite{lei2021less}      & 5.4M     & 37.4          & -             & 60.3          &     -         \\
ALPRO~\cite{li2022align}      & 5M     & 42.1          & 45.9             & -      &     -         \\
VIOLETv2~\cite{fu2022empirical}        & 5M     & 44.5          & \textbf{54.7}          & 72.8          &     -       \\

Clover~\cite{huang2022clover}        & 5M     & 43.9          & 51.9          & 71.4          &    -          \\
\rowcolor{mygray}
\textbf{COSA-B(Ours)}    & \textbf{5M}     & \textbf{46.2} &54.3& \textbf{73.4} &  \textbf{46.5} \\
\midrule

OmniVL~\cite{wang2022omnivl} &17M &44.1&51.0&-&-\\

HiTeA~\cite{ye2022hitea} & 17M &45.9&55.3&73.2&46.4 \\
SINGULARITY~\cite{lei2022revealing}   & 17M      & 43.5          &          -     &     -          & 43.1        \\

VINDLU-B~\cite{cheng2022vindlu} & 17M &43.8 &-&-&44.6\\ 
LAVENDER~\cite{li2022lavender}      & 30M      & 45.0          & \textbf{56.6}          & 73.5          &  -   \\
JustAsk~\cite{yang2021just}       & 69M      & 41.5          & 46.3          & -             & 38.9        \\

\rowcolor{mygray}
\textbf{COSA-B(Ours)}    & \textbf{17M}     &\textbf{46.9}  &55.5 & \textbf{75.0} & \textbf{47.3}  \\
\midrule
MERLOT~\cite{zellers2021merlot}        & 180M     & 43.1          & -             & 69.5          & 41.4        \\
All-in-one~\cite{wang2022all}    & 228.5M   & 46.8          & 48.3          & 66.3          &       -      \\

FrozenBiLM~\cite{yang2022zero}    & 410M     & 47.0          & 54.8          & 68.6          & 43.2         \\
mPLUG-2~\cite{xu2023mplug} & 417M &48.0&58.1&75.4&- \\
UMT-L~\cite{li2023unmasked} & 425M &47.1&55.2&-&47.9 \\ 

VALOR-L~\cite{chen2023valor}    & 433.5M & 49.2 & 60.0 & 78.7 & 48.6   \\

InternVideo~\cite{wang2022internvideo} &646M  & 47.1        &55.5 &72.2&- \\

GIT~\cite{wang2022git}      & 1.7B     & 43.2          & 56.8          & 72.8          &       -     \\

Flamingo(80B)~\cite{alayrac2022flamingo} & 2.3B     & 47.4          &          -     &        -       &   -         \\
VideoCoCa (2.1B)~\cite{yan2022video} &4.8B  & 46.0        &56.9 &-&- \\
   -      \\
GIT2 (5.1B)~\cite{wang2022git}         & 12.9B    & 45.6        & 58.2         & 74.9          &    -         \\

\rowcolor{mygray}
\textbf{COSA-L(Ours)}    & \textbf{417M}     &48.8 &  58.6 & 77.6 &49.2    \\

\rowcolor{mygray}
\textbf{COSA(Ours)(1.2B)}    &  \textbf{415M}   &\textbf{49.2} & \textbf{60.0} &\textbf{79.5}  &  \textbf{49.9}\\

\bottomrule
\end{tabular}}
\end{table*}

\begin{table}[]
\caption{Performance comparison on video captioning benchmarks. BLEU@4~\cite{papineni2002bleu} (B@4) and CIDEr~\cite{vedantam2015cider} (C) are used as metrics. Results of models utilizing additional modalities such as audio~\cite{chen2023valor} and subtitle~\cite{tang2021clip4caption++,li2021value} are denoted by dashed color. Following~\cite{tang2021clip4caption++,chen2023valor,wang2022git}, we employ SCST finetuning~\cite{rennie2017self} on the VATEX benchmark, and the corresponding results are marked with `*'. }
\label{sota_cap}
\centering
\scalebox{0.65}{
\begin{tabular}{ll|ll|cc|cc|cc|cc}
\toprule
\multirow{2}{*}{Method} & \multirow{2}{*}{Example} & \multicolumn{2}{c}{MSRVTT} & \multicolumn{2}{c}{MSVD} & \multicolumn{2}{c}{VATEX} & \multicolumn{2}{c}{YouCook2} & \multicolumn{2}{c}{TVC} \\

\cmidrule (lr){3-4} \cmidrule (lr){5-6} \cmidrule (lr){7-8} \cmidrule (lr){9-10} \cmidrule (lr){11-12}

                        &                          & B@4          & C           & B@4        & C           & B@4         & C           & B@4          & C            & B@4        & C          \\
\midrule

VIOLETv2~\cite{fu2022empirical}                & 5M                     & -            & 58.0        &    -        & 139.2       &      -       &   -          &        -      &          -    &      -      &       -     \\
HiTeA~\cite{ye2022hitea}                   & 5M                      &  -            & 62.5        &      -      & 145.1       &         -    &   -          &       -       &        -      &     -       &     -       \\
\rowcolor{mygray}
\textbf{COSA-B(Ours)}                 & \textbf{5M}                     &        \textbf{48.1}    & \textbf{64.3}           &   \textbf{66.9}       &  \textbf{146.0}         &     \textbf{40.5}     &     \textbf{72.6}      &   \textbf{7.8}         &   \textbf{100.9}          &      \textbf{15.6}      &      \textbf{59.2}    \\
\midrule

LAVENDER~\cite{li2022lavender}                & 30M                      & -            & 60.1        & -          & 150.7       &      -       &  -           &              &     -         &     -       &     -       \\

\rowcolor{mygray}
\textbf{COSA-B(Ours)}                 & \textbf{17M}                     &    \textbf{48.5}        &  \textbf{64.7}         &   \textbf{68.7}    &  \textbf{150.9}        &    \textbf{41.4}      &  \textbf{74.4}        &        \textbf{8.6}   & \textbf{106.2}          &     \textbf{16.2}       & \textbf{61.0}\\
\midrule
\textcolor{gray}{VALUE~\cite{li2021value}} & \textcolor{gray}{136M}                     &    -          &          -   &   -         &    -         & -           & \textcolor{gray}{58.1}        & \textcolor{gray}{12.4}         & \textcolor{gray}{130.3}        & \textcolor{gray}{11.6}       & \textcolor{gray}{50.5}       \\
\textcolor{gray}{CLIP4Caption++~\cite{tang2021clip4caption++}}          & \textcolor{gray}{400M}                     &          -    &      -       &         -   &           -  & \textcolor{gray}{40.6*}        & \textcolor{gray}{85.7*}        &       -       &    -          & \textcolor{gray}{15.0}       & \textcolor{gray}{66.0}       \\

\textcolor{gray}{VALOR-L~\cite{chen2023valor}}                   & \textcolor{gray}{433.5M}                   & \textcolor{gray}{54.4}         & \textcolor{gray}{74.0}        & \textcolor{gray}{80.7}       & \textcolor{gray}{178.5}       & \textcolor{gray}{45.6*}        & \textcolor{gray}{95.8*}        &          -    &    -          &   -         &       -     \\
GIT~\cite{wang2022git}                     & 1.7B                     & 53.8         & 73.9        & 79.5       & 180.2       & 41.6*        & 91.5*        & \textbf{10.3}         & 129.8        & 16.2       & 63.0       \\

Flamingo(80B)~\cite{alayrac2022flamingo}                & 2.3B                     &          -    &         -    &          -  &         -    &     -        &        -     & -            & 118.6        &   -         &    -        \\

GIT2 (5.1B)~\cite{wang2022git}             & 12.9B                    & \textbf{54.8}         & \textbf{75.9}        & \textbf{82.2}       & 185.4       & 42.7*        & 94.5*        & 9.4          & 131.2        & 16.9       & 66.1       \\

\rowcolor{mygray}
\textbf{COSA-L(Ours)}                   & \textbf{417M}                     &          53.2    &   72.1          &      74.0       &      169.1        & \textbf{45.4}    &  84.4           &      10.2        &     125.8         &          17.7  &  67.5          \\
\rowcolor{mygray}
\textbf{COSA(Ours)(1.2B)}      &                   \textbf{415M}                        &     53.7         & 74.7           &     76.5       &  178.5          &      43.7*       &  \textbf{96.5}*           &          10.1    &      \textbf{131.3}        &        \textbf{18.8}    & \textbf{70.7}    \\
\bottomrule
\end{tabular}}
\end{table}

\subsection{Comparison to State-of-the-Art}

\textbf{Text-to-Video Retrieval.}
We evaluate text-to-video retrieval on four benchmarks, including MSRVTT~\cite{xu2016msr}, DiDeMo~\cite{anne2017localizing}, LSMDC~\cite{rohrbach2017movie}, and ActivityNet~\cite{krishna2017dense}. ITC and ITM are used as the training objectives. During testing, we rank all candidates using similarity scores computed by ITC and then rerank the top-50 candidates via ITM. As shown in Table \ref{sota_ret}, COSA-B (5M) outperforms previous models of similar scale by a significant margin on all four benchmarks, especially for the long-form retrieval benchmark DiDeMo and ActivityNet. This demonstrates the effectiveness of concatenated sample training. It is noted that 
LF-VILA~\cite{sun2022long} is customized designed for long-form video retrieval with specialized framework and 
corpus, but fall behinds COSA-B which takes a simple unified architecture as well as common 
corpora. In addition, COSA (17M) evidently surpasses VINDLU~\cite{cheng2022vindlu} which have searched a best training recipe targeted 
at retrieval tasks. When compared to large-scale models, COSA achieves the best R@1 results on DiDeMo and ActivityNet datasets and comparable results on other benchmarks.

\textbf{Video QA.}
We evaluate video QA on four benchmarks, including MSRVTT-QA~\cite{xu2017video}, MSVD-QA~\cite{xu2017video}, TGIF-QA~\cite{li2016tgif}, and ActivityNet-QA~\cite{yu2019activitynet}, and  formulate it as an open-ended generative task to predict answers with questions as prefixes. As shown in Table \ref{sota_qa}, COSA-B (5M) and COSA-B (17M) outperform all models of similar scale by a significant margin. Under the large-scale pretraining scenario, COSA achieves new state-of-the-art accuracy on 4 benchmarks, using only 415M training data and 1.2B parameters.

\textbf{Video Captioning.}
We evaluate video captioning on five benchmarks, including MSRVTT~\cite{xu2016msr}, MSVD~\cite{chen2011collecting}, VATEX~\cite{wang2019vatex}, YouCook2~\cite{zhou2018towards}, and TVC~\cite{lei2020tvr}. The GM objective is used, and captions are autoregressively generated during inference. From the results in Table \ref{sota_cap}, we can see that COSA-B (5M) outperforms VIOLETv2, and HiTeA  which utilizes a more advanced vision backbone (MViTv2~\cite{fan2021multiscale}). COSA-B (17M) surpasses LAVENDER, which is trained on 30M vision-text pairs. When compared to large-scale foundation models, COSA achieves comparable performances on the MSRVTT and MSVD datasets. Additionally, COSA outperforms GIT2, which specializes in captioning tasks, and achieves new state-of-the-art performances on both open-domain benchmarks~\cite{wang2019vatex} and domain benchmarks~\cite{zhou2018towards,lei2020tvr} (cooking and TV), using only 24\% of its parameters and 3.2\% of the training examples. It is worth noting that both GIT2 and COSA do not use any video-text corpus in the pretraining stage, but COSA  can online generation of pseudo long-form videos. Furthermore, utilizing the vision modality alone, COSA achieves  higher CIDEr scores than VALOR and CLIP4Caption++, which additionally incorporate audio and subtitle modalities.

\textbf{Image-Text Benchmarks.}
We evaluate COSA on image-text benchmarks, including text-to-image retrieval on MSCOCO~\cite{lin2014microsoft} and Flickr30K~\cite{plummer2015flickr30k} (zero-shot), captioning on MSCOCO, and image QA on VQAv2~\cite{goyal2017making}. As shown in Table \ref{sota_img}, COSA surpasses BLIP-2 and BEiT-3, and achieves new state-of-the-art results on the MSCOCO and Flickr30K benchmarks. Additionally, COSA achieves the best SPICE score (27.0) on MSCOCO caption benchmark and comparable results on  VQAv2 benchmark.

\begin{table}[]
\caption{Performance comparison on text-to-image retrieval, image captioning, and image QA benchmarks. `ZS' stands for zero-shot evaluation. CIDEr (C) and SPICE (S) are reported for captioning. SCST finetuning  is employed on COCO caption, and corresponding results are marked with `*'.}
\label{sota_img}
\centering
\scalebox{0.65}{
\begin{tabular}{ll|ccc|ccc|cc|cc}
\toprule
                               &                           & \multicolumn{3}{c}{MSCOCO-Ret} & \multicolumn{3}{c}{Flickr30K-Ret (ZS)} & \multicolumn{2}{c}{MSCOCO-Cap} & \multicolumn{2}{c}{VQAv2} \\
                             \cmidrule (lr){3-5} \cmidrule (lr){6-8} \cmidrule (lr){9-10} \cmidrule (lr){11-12} \cmidrule (lr){11-12}
\multirow{-2}{*}{Method}       & \multirow{-2}{*}{Example} & R@1      & R@5      & R@10     & R@1       & R@5       & R@10      & C              & S             & dev         & std         \\
\midrule
OFA~\cite{wang2022ofa}    &  18M                         & -        & -        & -        & -         & -         & -         & \textbf{154.9}*          & 26.6*          & 82.0        & 82.0        \\
BEiT-3~\cite{wang2022image}  & 21M                       & 67.2     & 87.7     & 92.8     & 81.5     & 95.6      & 97.8       & 147.6          & 25.4          & \textbf{84.19}       & \textbf{84.03}       \\

BLIP~\cite{li2022blip}                           & 129M                      & 65.1     & 86.3     & 91.8     & 86.7      & 97.3      & 98.7      & 136.7          & -             & 78.25       & 78.32       \\
BLIP-2~\cite{li2023blip} & 129M                      & 68.3     & 87.7     & 92.6     & 89.7          &  98.1         &  98.9          & 145.8          & -             & 82.19       & 82.30       \\
mPLUG-2~\cite{xu2023mplug}                        & 417M                       & 65.7     & 87.1     & 92.6     & -     & -      & -      & 137.7          & 23.7          & 81.11       & 81.13       \\
VALOR~\cite{chen2023valor}                          & 433.5M                     & 61.4     & 84.4     & 90.9     & -         & -         & -         & 152.5*          & 25.7*          & 78.46       & 78.62       \\
Florence~\cite{yuan2021florence}                       & 900M                      & 63.2     & 85.7     & -        & 76.7    &93.6      & -         & -              & -             & 80.16       & 80.36       \\
GIT~\cite{wang2022git}                            & 1.7B                      & -        & -        & -        & -         & -         & -         & 151.1*          & 26.3*          & 78.6        & 78.8        \\

SimVLM~\cite{wang2021simvlm}                         & 1.8B                      & -        & -        & -        & -         & -         & -         & 143.3          & 25.4          & 80.03       & 80.34       \\

ALIGN~\cite{jia2021scaling}                          & 1.8B                      & 59.9     & 83.3     & 89.8     & 75.7      & 93.8      & 96.8      & -              & -             & -           & -           \\

Flamingo (80B)]~\cite{alayrac2022flamingo}             & 2.3B                      & -        & -        & -        & -         & -         & -         & 138.1          & -             & 82.0        & 82.1        \\
CoCa(2.1B)~\cite{yu2022coca}                           & 4.8B                      & -        & -        & -        & -         & -         & -         & 143.6          & 24.7          & 82.3        & 82.3        \\

GIT2(5.1B)~\cite{wang2022git}                           & 12.9B                     & -        & -        & -        & -         & -         & -         & 152.7*          & 26.4*          & 81.7        & 81.9        \\

\rowcolor{mygray}
\textbf{COSA(Ours)(1.2B)}                           &    \textbf{415M}                       &    \textbf{68.5}       &    \textbf{88.0}       &  \textbf{93.0}        &      \textbf{90.2}      &      \textbf{98.4}      &   \textbf{99.4}        & 150.6*               &     \textbf{27.0*}          &         80.46    &  80.54           \\
\bottomrule
\end{tabular}}
\end{table}

\begin{table}[]
\caption{Experiment results of SST (baseline), COSA, and COSA variants on a broad range of video/image-language benchmarks. R@1, CIDEr, and Acc are reported for retrieval, captioning, and QA tasks, respectively.}
\centering
\label{effect of COSA}
\scalebox{0.75}{
\begin{tabular}{l|cccccccccccc}
\toprule
\multirow{2}{*}{Exp} & \multicolumn{4}{c}{Video Ret}      & \multicolumn{3}{c}{Video Cap} & \multicolumn{2}{c}{Video QA} & Image Ret & \multicolumn{2}{c}{Image Cap} \\ \cmidrule (lr){2-5} \cmidrule (lr){6-8} \cmidrule (lr){9-10} \cmidrule (lr){11-11} \cmidrule (lr){12-13}
                     & \rotatebox{90}{DiDeMo} & \rotatebox{90}{ActivityNet} & \rotatebox{90}{MSRVTT} &  \rotatebox{90}{VATEX} & \rotatebox{90}{YouCook}  & \rotatebox{90}{VATEX} & \rotatebox{90}{TVC} & \rotatebox{90}{MSRVTT}       & \rotatebox{90}{TGIF-QA}      & \rotatebox{90}{MSCOCO}  & \rotatebox{90}{MSCOCO}        & \rotatebox{90}{VIST}        \\ 
\midrule
SST               & 49.2    & 43.8     & 39.4    & 62.4      & 85.4        & 67.6     & 53.9  & 45.9        & 73.0        & 53.9      & 124.7           & 14.8       \\
COSA                & \textbf{54.8}    & \textbf{51.2}     & \textbf{43.5}    & \textbf{63.9}      & \textbf{91.9}        & \textbf{68.1}     & \textbf{55.4}  & \textbf{46.5}        & \textbf{73.7}         & \textbf{54.7}      & \textbf{125.4}          & \textbf{20.7}       \\
COSA-copy                    & 43.5    & 39.3     & 42.3    & 60.8      & 84.6        &       65.8     & 52.0       & 45.4        &     72.2
        &       51.6
     &       122.7          & 11.9       \\
COSA-shuffle                    & 50.9    & 48.1     & 44.7    &64.0           & 88.8         &     67.7
    &    54.4      & 46.6        &     73.9

         &      54.5

  &        124.9            & 14.2       \\
\bottomrule
\end{tabular}}
\end{table}

\begin{table}[]
\caption{Experiment results of different training objectives for COSA. R, C, Q in the parentheses represent retrieval, captioning, and QA tasks, respectively.}
\centering
\label{loss}
\scalebox{0.75}{
\begin{tabular}{lllllllll|lllll}

\toprule
&ITC & ITM & MLM & GM & CITC & CITM & CMLM & CGM & DiD(R) & ANET(R) & MSR(R) & TVC(C) & TGIF(Q) \\
\midrule
a&\checkmark   & \checkmark    & \checkmark    & \checkmark   &      &      &      &     & 49.2   & 43.8    & 39.4   & 53.9   & 73.0    \\
b&\checkmark    & \checkmark    &     &    &      &      & \checkmark     & \checkmark    & 53.0   & 50.6    & 42.8   & \textbf{55.7}   & 73.7    \\
c&    &     &     &    & \checkmark     & \checkmark     & \checkmark     & \checkmark    & 52.3   & 49.9    & 42.3   & 55.5   & 73.6    \\
d&\checkmark    & \checkmark    &     &    & \checkmark     & \checkmark     & \checkmark     & \checkmark    & \textbf{54.8}   & \textbf{51.2}    & \textbf{43.5}   & 55.4   & \textbf{73.7}    \\
e&\checkmark    & \checkmark    & \checkmark    & \checkmark   & \checkmark     & \checkmark     & \checkmark     & \checkmark    & 52.3   & 49.9    & 42.3   & 55.5   & 73.6 \\
\bottomrule
\end{tabular}}
\end{table}

\begin{table}[]
\caption{Experiment results of changing dataset and vision backbone (Enc).}
\centering
\label{robustness}
\scalebox{0.75}{
\begin{tabular}{llll|lllllll}
\toprule
Exp& Enc & Data    & \#Frame  & DiD(R) & ANET(R) & MSR(R) & YOU(C) & VIST(C) & TVC(C) & MSR(Q) \\
\midrule
SST& Swin-B   & CC3M       & -              & 35.3   & 30.4    & 34.7       & 91.2   & 14.3    & 51.6   & 44.6   \\
COSA&Swin-B   & CC3M       & -           & \textbf{45.1}   & \textbf{41.7}    & \textbf{36.2}       & \textbf{94.6}   & \textbf{22.3}    & \textbf{53.1}   & \textbf{44.9}   \\
\midrule
SST&CLIP-B   & WebVid2.5M & 1              & 43.6   & 40.2    & 40.7       & 89.3   & 12.2    & 53.5   & 46.0   \\
SST&CLIP-B   & WebVid2.5M & 4            & 51.4   & 47.3    & 42.5       & 90.8   & 11.9    & 54.0   & 46.2   \\
COSA&CLIP-B   & WebVid2.5M & 1          & \textbf{54.7}   & \textbf{52.2}    & \textbf{44.0}       & \textbf{93.4}    & \textbf{17.4}    & \textbf{55.0}   & \textbf{46.4}   \\
\bottomrule
\end{tabular}}
\end{table}

\subsection{Ablation Study}
In this subsection, we conduct extensive ablation experiments to demonstrate the effect of sample concatenation  and the method's robustness to different vision backbones and training corpora. We also explore the influence of training objectives, iterations, concatenation numbers, and sampling methods. Besides above mentioned benchmarks, we further evaluate the model on the Visual Storytelling Benchmark (VIST), which requires models to perceive the relationships among multiple images and describe them in sequential paragraphs, thus demanding strong narrative abilities. Visualizations of COSA’s prediction on VIST can be found in Appendix.

 \begin{figure*}[h]
\centering
\includegraphics[width=0.9\linewidth]{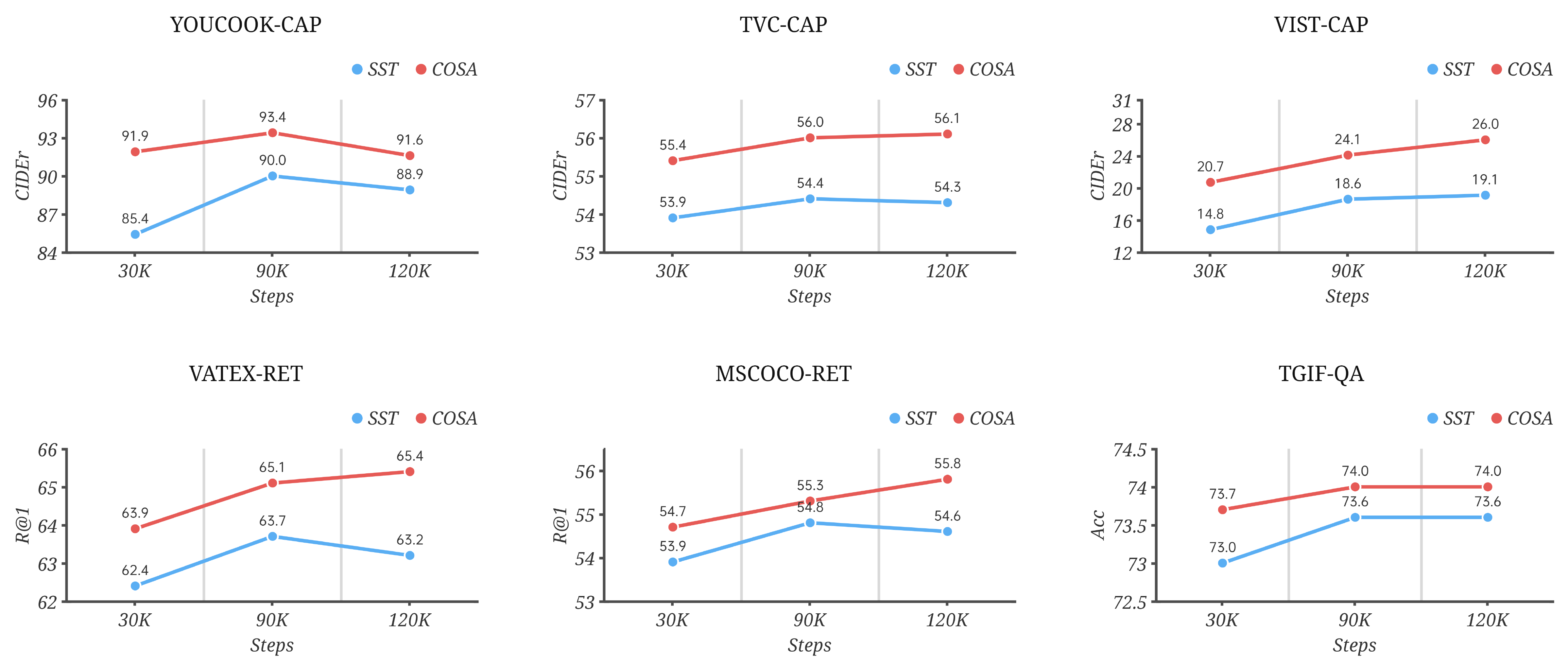}
\caption{Experiment results of SST and COSA when training for more iterations.}
\label{fig:iters}
\end{figure*}

\textbf{Effect of COSA.} As presented in Table \ref{effect of COSA}, COSA consistently improves performance on 12 benchmarks, including  long-form (DiDeMo and ActivityNet) and short-form video-language as well as image-text tasks, with significant margins compared to the traditional Single Sample Training baseline (SST). We further investigate two COSA variants: COSA-copy and COSA-shuffle. COSA-copy simply duplicates a single sample (image and text) multiple times and concatenates them, instead of performing cross-sample concatenation as in COSA. On the other hand, COSA-shuffle concatenates cross-samples from a batch, but with the captions randomly shuffled before concatenation to disrupt event-level temporal correspondence. The results indicate that the improvements brought by COSA primarily arise from cross-sample concatenation, as the performance drops drastically below the baseline when this operation is omitted. The inclusion of multiple images and longer sentences alone does not yield the same improvements. Moreover, preserving the concatenation order encourages models to learn explicit scene-level temporal correspondence, further enhancing performance on various benchmarks, such as long-form video retrieval, video captioning, and VIST.

\begin{table}[t]
\caption{Experiment results of different sampling methods for online sample concatenation.}
\centering
\label{similarity}
\scalebox{0.8}{
\begin{tabular}{l|lllllll}
\toprule
Sampling                                                  & DiD(R) & ANet(R) & MSR(R) & VAT(R) & TVC(C) & VIST(C) & MSR(Q) \\
\midrule
random                                                    & \textbf{54.8}   & \textbf{51.2}    & \textbf{43.5}   & \textbf{63.9}   &  \textbf{55.4}      & \textbf{20.7}    & \textbf{46.5}   \\
vision similarity                                         & 53.7   & 50.9    &   43.0     & 63.6   &  55.1      & 20.4    & 46.4   \\
text similarity  & 48.2   & 45.4    & 42.6   & 62.4   &  55.2      & 20.2    & 46.3  \\
\bottomrule
\end{tabular}
}
\end{table}

 \begin{figure*}[t]
\centering
\includegraphics[width=0.9\linewidth]{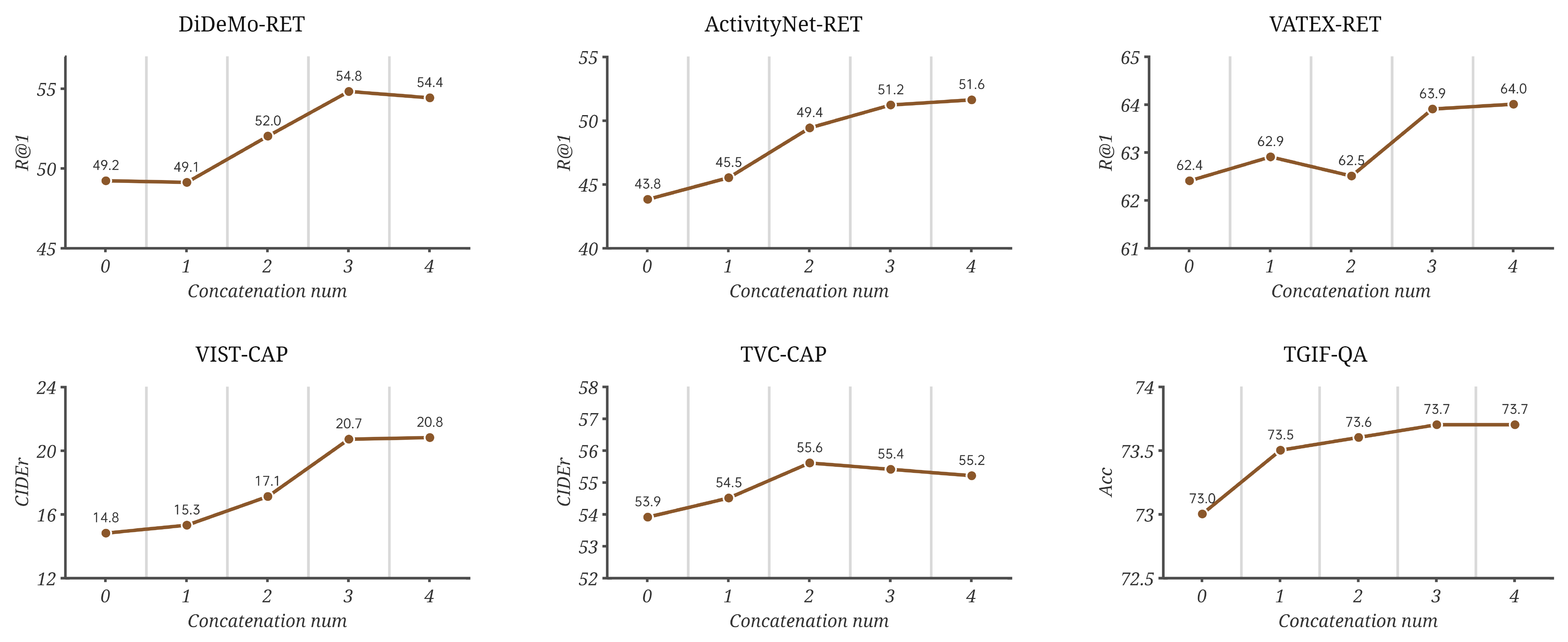}
\caption{Experiment results of different concatenation numbers used in COSA.}
\label{fig:concate num}
\end{figure*}

\textbf{Training Objectives.} As depicted in Table \ref{loss}, compared to the SST baseline, incorporating COSA while modifying the  MLM and GM objectives to CMLM and CGM leads to evident improvements across all five benchmarks (b vs a). However, directly replacing the ITC and  ITM objectives with to CITC and CITM  results in performance degradation on all benchmarks (c vs b). The default objectives (d) yield the best results for most benchmarks.

\textbf{Robustness.} We further explore the robustness of COSA with respect to different vision encoders and training corpora. The previous experiments employed CLIP-B as the vision backbone and were conducted on the CC3M dataset. In this experiment, we replace them with Swin Transformer~\cite{liu2021swin} and WebVid2.5M, respectively. From the results in Table \ref{robustness}, it is evident that COSA consistently outperforms SST under different backbone and corpus settings. Additionally, contrary to the conclusion of Singularity~\cite{lei2022revealing}, we observe that sampling multiple frames can indeed improve downstream performance (except for VIST), but is still much less effective when compared with proposed COSA.

\textbf{More Iterations.} In the previous experiments, models were trained for 30K steps for efficiency purposes. One might suspect that the improvements of COSA could be attributed to faster convergence rather than cross-sample learning and event-level temporal correspondence, as the model encounters each sample multiple times per iteration. However, the results in Figure \ref{fig:iters} demonstrate that as training iterations prolong, the performance gap between COSA and SST remains consistent, even when SST has already converged. Moreover, pretraining with COSA for 30K steps (10 hours) outperforms pretraining with SST for 120K steps (23 hours) on all benchmarks.

\textbf{Different Concatenate Number.} Figure \ref{fig:concate num} illustrates the performance of models trained with different concatenate numbers ($n_c$) ranging from 0 (SST baseline) to 4. Based on the results, we choose $n_c=3$ as it consistently performs well across all benchmarks.

\textbf{Sampling Choice.} By default, we randomly choose samples within a batch for grouping and concatenation. We also experimented with two alternative choices: selecting the top $n_c$ most similar samples based on the similarity matrix computed by either the image or text encoder output features. The results in Table \ref{similarity} demonstrate the importance of reducing semantic relevance among grouped samples through random sampling.

\section{Conclusion and Limitation}
This paper introduces COSA, a concatenated sample pretrained   unified vision-language foundation model. COSA transforms image-text corpora into pseudo long-form video-text corpora, enabling explicit event-sentence correspondence and promoting event-level temporal learning in foundation models. Extensive ablation studies demonstrate the effectiveness of COSA in improving various downstream tasks, including long-form/short-form video-language tasks and image-text tasks. COSA achieves state-of-the-art results on multiple benchmarks. However, it is important to note that the proposed COSA has only been proven effective in the vision-language pretraining field, and we hope that it can inspire further research on co-learning multiple samples in other AI communities.

\section*{Appendix}
\appendix

\section{Pretraining Settings}

\begin{table}[h]
\caption{Pretraining Settings for COSA models. '*' means that captions in those corpora has been replaced by descriptions generated by a separately trained captioner.}
\label{pretrain settings}
\centering
\scalebox{0.9}{
\begin{tabular}{llllllll}
\toprule
Model                   & Vision Encoder                 & param                 & Sample                     & Training Corpus & Batch Size & Steps & Epoch \\

\midrule

\multirow{2}{*}{COSA-B} & \multirow{2}{*}{Swin-B}        & \multirow{2}{*}{251M} & \multirow{2}{*}{5M}        & CC3M            & 2048       & 50000 & 35    \\
                        &                                &                       &                            & WebVid2.5M      & 2048       & 50000 & 41    \\

\midrule
\multirow{3}{*}{COSA-B} & \multirow{3}{*}{Swin-B}        & \multirow{3}{*}{251M} & \multirow{3}{*}{17M}       & CC4M            & 2048       & 60000 & 25    \\
                        &                                &                       &                            & CC12M           & 2048       & 60000 & 12    \\
                        &                                &                       &                            & WebVid2.5M      & 2048       & 40000 & 33    \\
\midrule
\multirow{3}{*}{COSA-L} & \multirow{3}{*}{CLIP-ViT-L}    & \multirow{3}{*}{468M} & \multirow{3}{*}{(17+400)M} & CC4M            & 2048       & 60000 & 25    \\
                        &                                &                       &                            & CC12M           & 2048       & 60000 & 12    \\
                        &                                &                       &                            & WebVid2.5M      & 2048       & 35000 & 29    \\
\midrule
\multirow{3}{*}{COSA}   & \multirow{3}{*}{EVAClip-ViT-G} & \multirow{3}{*}{1.2B} & \multirow{3}{*}{(15+400)M} & CC4M            & 2048       & 50000 & 20    \\
                        &                                &                       &                            & CC12M*           & 2048       & 50000 & 10    \\
                        &                                &                       &                            & LAION-102M*      & 2048       & 50000 & 1    \\
\bottomrule
\end{tabular}}
\end{table}

The specific pretraining settings of all scales of COSA models are presented in Table \ref{pretrain settings}. It is noted that the training corpus of the largest one (COSA) has been clean by a separately trained captioner. Specifically, the captioner takes the same model architecture of COSA and is pretrained with generative modeling loss (GM) only on the same corpus with original captions (CC4M, CC12M and LAION-102M) without concatenated samples applied. After pretraining, the captioner is finetuned with GM  loss on a mixture of downstream of datasets including MSRVTT, VATEX, MSVD and MSCOCO. At last we use trained captioner to generate high-quality captions for CC12M and LAION-102M with top-10 sampling method. 
\section{Downstream  Dataset Descriptions}
COSA is evaluated on a series of downstream datasets  including MSRVTT, MSVD, LSMDC, DiDeMo, VATEX, YouCook2,    ActivityNet Caption, TGIF, TVC, MSCOCO, Flickr30K and VQAv2.  Train/val/test splits of different benchmarks of those datasets are presented in Table \ref{split}. 

Specifically, \textbf{MSRVTT}~~\cite{xu2016msr} dataset consists of 10K video clips and 20K captions. Our evaluation encompasses text-to-video retrieval, video captioning, and video QA tasks on this dataset. For retrieval, we adopt the `1K-A split' and  for captioning and QA, we adhere to the standard split. \textbf{MSVD}~\cite{chen2011collecting} dataset comprises 1,970 videos, each accompanied by approximately 40 captions. Our evaluation involves video captioning using the official split and video QA using the split proposed by Xu et al.~\cite{xu2017video}. \textbf{LSMDC}~~\cite{rohrbach2017movie} dataset consists of 118,000 clips from 202 movies, with each clip paired with a single caption. We evaluate text-to-video retrieval on this dataset using the official split. \textbf{DiDeMo}~~\cite{anne2017localizing} dataset includes 10K long-form videos sourced from Flickr. For each video, four short sentences are annotated in temporal order. We concatenate these short sentences and evaluate `paragraph-to-video' retrieval and the official split is employed. \textbf{VATEX}~\cite{wang2019vatex} dataset contains 41,250 video clips sourced from the Kinetics-600 dataset~\cite{kay2017kinetics}, accompanied by 825,000 sentence-level descriptions. Our evaluation focuses on text-to-video retrieval and video captioning tasks. Regarding captioning, we employ the official split, while for retrieval, we follow the HGR~~\cite{chen2020fine} split protocol. \textbf{YouCook2}~~\cite{zhou2018towards} dataset comprises 14K video clips extracted from 2K instructional cooking videos on YouTube. Each video showcases multiple actions performed by a chef, along with corresponding textual descriptions and temporal annotations. We assess  video captioning on this dataset using the official splits. \textbf{ActivityNet Caption}~\cite{krishna2017dense} dataset encompasses 20K long-form videos sourced from YouTube whose average length are  180 seconds. We evaluate text-to-video retrieval and video QA on this dataset. For retrieval, we use the official split, while for video QA, we adopt the split proposed by Yu et al.~\cite{yu2019activitynet}. \textbf{TGIF}~~\cite{jang2017tgif} dataset features three video QA benchmarks, including TGIF-Action, TGIF-Transition, and TGIF-Frame, with the first two being multiple-choice QA and the last one being open-ended QA. We evaluate VAST on the TGIF-Frame benchmark using the official split. \textbf{TVC}~~\cite{lei2020tvr} dataset is a multi-channel video captioning dataset, comprising 108,000 video moments and 262,000 paired captions. Additional input in the form of video subtitles can be utilized. We evaluate video captioning on this benchmark using the official split with vision information utilized only. \textbf{MSCOCO}~~\cite{lin2014microsoft} dataset contains 123K images, each paired with  five annotated captions. We evaluate text-to-image retrieval and image captioning tasks on this dataset using the Karpathy split~~\cite{karpathy2015deep}. \textbf{Flickr30K}~~\cite{plummer2015flickr30k} dataset consists of 31K images, with each image paired with five annotated captions. We assess text-to-image retrieval on this dataset using the Karpathy split~~\cite{karpathy2015deep}. \textbf{VQAv2}~~\cite{goyal2017making} dataset  comprises 1.1 million questions and 11.1 million answers related to MSCOCO images and we utilize train+val set for training and evaluate on the online test server.

\begin{table}[]
\caption{Downstream datasets splits. }
\label{split}
\centering
\scalebox{0.85}{
\begin{tabular}{llllllll}
\toprule
\multirow{2}{*}{Task}                    & \multirow{2}{*}{Benchmark} & \multicolumn{3}{c}{\#Videos/\#Images} & \multicolumn{3}{c}{\#Captions/\#QA-pairs} \\
\cmidrule(lr){3-5}  \cmidrule(lr){6-8}
                                         &                            & Train     & Val    & Test         & Train    & Val      & Test             \\
\midrule
\multirow{4}{*}{Text-to-Video Retrieval} & MSRVTT                     & 9000      & -      & 1000         & 180000   & -        & 1000             \\
                                         & DiDeMo                     & 8394      & 1065   & 1003         & 8394     & 1065     & 1003             \\
                                         & ANET                       & 10009     & -      & 4917         & 10009    & -        & 4917             \\
                                         & LSMDC                      & 101046    & 7408   & 1000         & 101046   & 7408     & 1000             \\
\midrule
\multirow{5}{*}{Video Captioning}        & MSRVTT                     & 6513      & 497    & 2990         & 130260   & 9940     & 59800            \\
                                         & YouCook2                   & 10337     & 3492   & -            & 10337    & 3492     & -                \\
                                         & MSVD                       & 1200      & 100    & 670          & 48774    & 4290     & 27763            \\
                                         & VATEX                      & 25991     & 3000   & 6000         & 259910   & 30000    & 60000            \\
                                         & TVC                        & 86603     & 10841  & -            & 174350   & 43580    & -                \\
\midrule
\multirow{4}{*}{Video QA}                & MSVD-QA                    & 1200      & 250    & 520          & 30933    & 6,415    & 13157            \\
                                         & TGIF-FrameQA               & 32345     & -      & 7132         & 39389    & -        & 13691            \\
                                         & MSRVTT-QA                  & 6513      & 497    & 2990         & 158581   & 12278    & 72821            \\
                                         & ANET-QA                    & 3200      & 1800   & 800          & 32000    & 18000    & 8000             \\
\midrule
\multirow{2}{*}{Text-to-Image retrieval} & MSCOCO                     & 113287   & 5000  & 5000        & 566747   & 25010    & 25010            \\
                                         & Flickr30K                  & 29000     & 1014   & 1000         & 145000   & 5070     & 5000             \\
\midrule
Image Captioning                         & MSCOCO                     & 113287   & 5000  & 5000        & 566747   & 25010    & 25010            \\
\midrule
Image QA                                 & VQAv2                      & 82783     & 40504  & 37K/81K  & 4437570  & 2143540  & 1.1M/4.5M \\
\bottomrule
\end{tabular}}
\end{table}

\begin{table}[]
\caption{Downstream task finetuning settings of COSA models. Lr, Bs, Epo, F\_train, F\_test and Res denote learning rate, batch size, epoch, training sampling frames, testing sampling frames and resolution, respectively. }
\label{finetune settings}
\centering
\scalebox{0.9}{

\begin{tabular}{lllllccl}
\toprule
Task                              & Benchmark    & Lr     & Bs  & Epo & F\_train & F\_test & Res \\
\midrule
\multirow{5}{*}{Text-to-Video Retrieval}                 & MSRVTT       & 2e-5   & 64  & 3.6   & 8         & 16       & 224 \\
                                  & VATEX        & 2e-5   & 64  & 2.5   & 8         & 16       & 224 \\
                                  & DiDeMo       & 2e-5   & 64  & 40    & 8         & 32       & 224 \\
                                  & ANET         & 2e-5   & 64  & 20    & 8         & 32       & 224 \\
                                  & LSMDC        & 2e-5   & 64  & 5     & 8         & 32       & 224 \\
\midrule
\multirow{7}{*}{Video Captioning} & MSRVTT       & 2e-5   & 128 & 10    & 8         & 8        & 224 \\
                                  & YouCook2     & 3e-5   & 64  & 30    & 8         & 16       & 224 \\
                                  & MSVD         & 1e-5   & 64  & 1.2   & 8         & 8        & 224 \\
                                  & VATEX        & 2e-5   & 64  & 10    & 8         & 20       & 224 \\
                                  & VATEX(SCST)  & 7e-6   & 64  & 5     & 8         & 20       & 224 \\
                                  & TVC          & 3e-5  & 64  & 40    & 8         & 8        & 224 \\
\midrule
\multirow{4}{*}{Video QA}         & MSVD-QA         & 1e-5   & 64  & 8     & 8         & 18       & 224 \\
                                  & TGIF-FrameQA        & 2e-5   & 64  & 10    & 4         & 4        & 224 \\
                                  & MSRVTT-QA       & 2e-5   & 64  & 4.5   & 8         & 8        & 224 \\
                                  & ANET-QA         & 2e-5   & 64  & 10    & 8         & 16       & 224 \\
\midrule
Text-to-Image retrieval           & MSCOCO       & 1e-5   & 256 & 5     & -         & -        & 384 \\
\midrule
\multirow{2}{*}{Image Captioning} & MSCOCO       & 1e-5   & 64  & 5     & -         & -        & 480 \\
                                  & MSCOCO(SCST) & 2.5e-6 & 64  & 2.5   & -         & -        & 480 \\
\midrule
Image QA                          & VQAv2        & 2e-5     & 128 & 20    & -         & -        & 384 \\
\bottomrule
\end{tabular}}
\end{table}
\section{Finetuning Settings}
Specific finetuning settings of COSA model such as learning rate, batch size, training epochs and sampled frames are presented in Table \ref{finetune settings}.

\section{Additional Results}
\subsection{Text-to-Video Retrieval on VATEX Benchmark}
As shown in Table \ref{vatex ret}, COSA-B which takes Swin-B as vision backbone has already outperformed the performance of CLIP4Clip that takes CLIP-ViT-B as backbone. In addition, COSA evidently surpass VALOR~~\cite{chen2023valor} which use additional audio modality as input, with less training data. These results further demonstrate the effectiveness of proposed method.  

\subsection{Qualitative Comparison on VIST Benchmark}
\begin{figure*}[]
\centering
\includegraphics[width=0.85\linewidth]{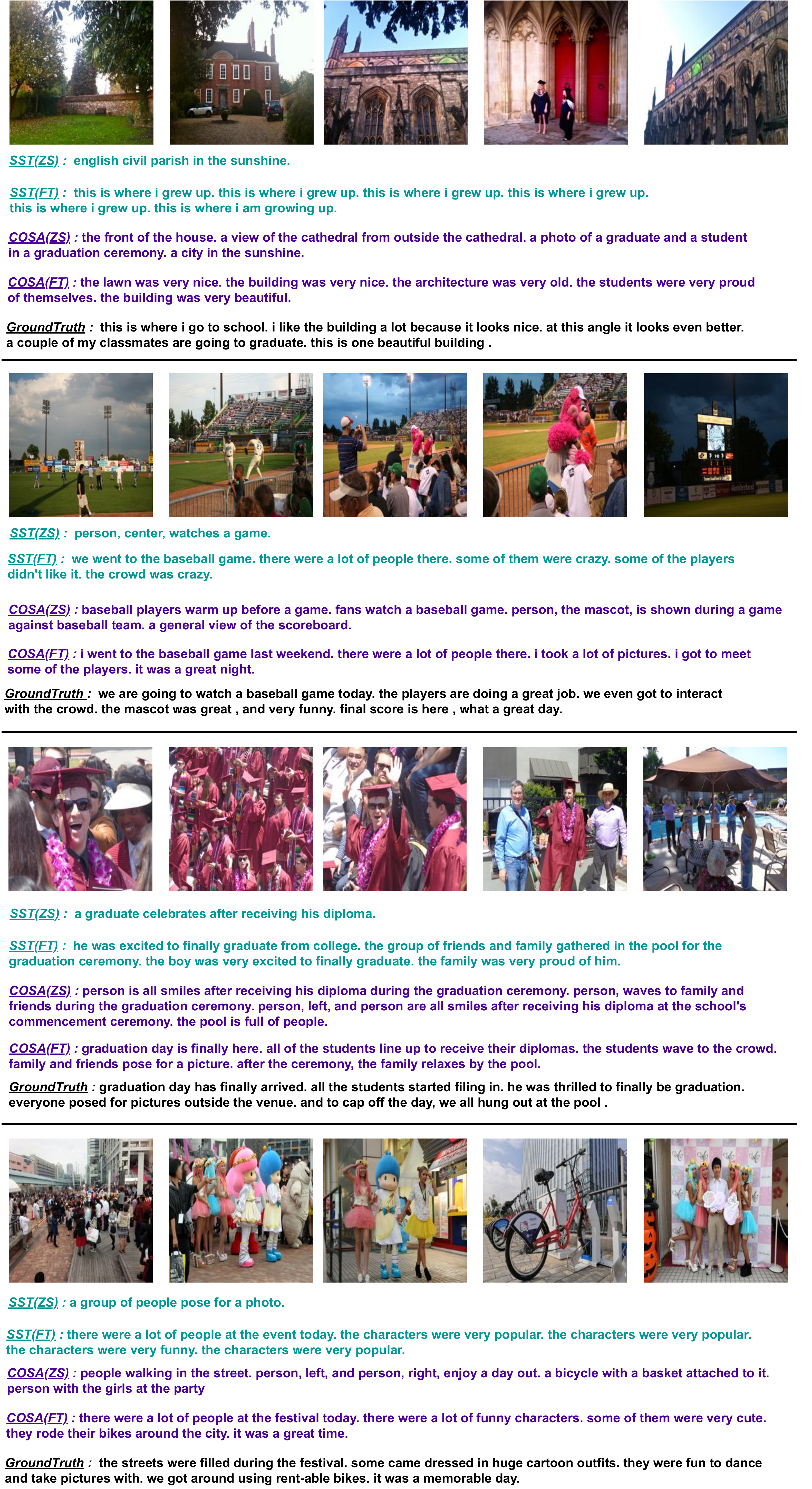}
\caption{Qualitative comparison between SST and COSA on the visual story telling (VIST) benchmark. ZS and FT denote zero-shot and finetuning, respectively.}
\label{fig:vist}
\end{figure*}

The qualitative comparisons between model trained with SST (Single Sample Training) and proposed COSA (Concatenated Sample Training) prediction on visual story telling benchmark (VIST) is shown in Figure \ref{fig:vist}, from which we can find that COSA can generate comprehensive captions (storys) for multiple frames under both zero-shot and finetuning settings, while SST can only generate one sentence describing limited contents under zero-shot setting, and is easy to generate duplicated captions under finetuning settings.

\begin{table}[]
\caption{Results Comparison on VATEX text-to-video retrieval benchmark. }
\label{vatex ret}
\centering
\scalebox{1.0}{
\begin{tabular}{lllll}
\toprule
Method      & Example & R@1  & R@5  & R@10 \\
\midrule
Support-Set~\cite{patrick2020support} & 136M    & 44.9 & 82.1 & 89.7 \\
CLIP4Clip~\cite{luo2022clip4clip}   & 400M    & 55.9 & 89.2 & 95.0 \\
DCR~\cite{wang2022disentangled}         & 400M    & 65.7 & 92.6 & 96.7 \\
VALOR~\cite{chen2023valor}       & 433.5M  & 76.9 & 96.7 & 98.6 \\
\midrule
\rowcolor{mygray}
\textbf{COSA-B}      & \textbf{5M}      & 65.3 & 92.0 & 96.0 \\
\rowcolor{mygray}
\textbf{COSA-B}      & \textbf{17M}     & 67.7 & 93.4 & 96.9 \\
\rowcolor{mygray}
\textbf{COSA-L}      & \textbf{417M}    & 75.4 & 95.9 & 98.2 \\
\rowcolor{mygray}
\textbf{COSA}        & \textbf{415M}    & \textbf{78.6} & \textbf{97.1} & \textbf{98.7} \\
\midrule
\end{tabular}}
\end{table}

\newpage
\normalem
{
\small
\bibliographystyle{IEEEtran}
\bibliography{cosa}
}

\end{document}